\pdfoutput=1

\documentclass[11pt]{article}

\usepackage{acl}

\usepackage{times}
\usepackage{latexsym}

\usepackage[T1]{fontenc}

\usepackage[utf8]{inputenc}

\usepackage{microtype}

\usepackage{amsmath}
\usepackage{amssymb}
\usepackage{graphicx}
\usepackage{tikz}
\usepackage{pgf}
\usepackage{xcolor}
\usepackage{booktabs}
\usepackage{enumitem}
\usepackage{subcaption}

\usepackage{contour}
\usepackage[normalem]{ulem}

\contourlength{1pt}
\newcommand{\myuline}[1]{%
  \uline{\phantom{#1}}%
  \llap{\contour{white}{#1}}%
}

\usepackage[hang]{footmisc}
\title{Multivalent Entailment Graphs for Question Answering}

\author{Nick M{\raisebox{0.3ex}{c}}Kenna$^{\dag}$ \quad Liane Guillou$^{\dag}$ \quad Mohammad Javad Hosseini$^{\dag\ddag}$\thanks{Now at Google Research.} \\ {\bf Sander Bijl de Vroe}$^{\dag}$ \quad {\bf Mark Johnson}$^\S$ \quad {\bf Mark Steedman}$^{\dag}$ \\ $^{\dag}$University of Edinburgh \quad $^{\ddag}$Alan Turing Institute, UK \quad $^\S$Oracle Digital Assistant \\ \texttt{\{nick.mckenna, javad.hosseini, sbdv\}@ed.ac.uk} \\ \texttt{\{lguillou, steedman\}@inf.ed.ac.uk} \\ \texttt{mark.mj.johnson@oracle.com}}
        
\begin{document}
\maketitle
\begin{abstract}
Drawing inferences between open-domain natural language predicates is a necessity for true language understanding. There has been much progress in unsupervised learning of entailment graphs for this purpose. We make three contributions: (1) we reinterpret the Distributional Inclusion Hypothesis to model entailment between predicates of different valencies, like $\textsc{defeat}$(Biden, Trump) $\vDash \textsc{win}$(Biden); (2) we actualize this theory by learning unsupervised \textit{Multivalent Entailment Graphs} of open-domain predicates; and (3) we demonstrate the capabilities of these graphs on a novel question answering task. We show that directional entailment is more helpful for inference than non-directional similarity on questions of fine-grained semantics. We also show that drawing on evidence across valencies answers more questions than by using only the same valency evidence.
\end{abstract}

\section{Introduction}
We are reading a mystery about a dark and foreboding manor and have one question: ``is Mr. Boddy dead?''\footnote{The murder mystery board game \textit{Clue} (also known as \textit{Cluedo}) lends inspiration to this project.} Our text might say ``Colonel Mustard killed Mr. Boddy,'' or ``Mr. Boddy was murdered in the kitchen with a candlestick,'' either of which answers the question, but only via natural language inference. An \textit{Entailment Graph} (EG) is a structure of meaning postulates supporting these inferences such as ``if A kills B, then B is dead.''

Entailment Graphs contain vertices of open-domain natural language predicates and entailments between them are represented as directed edges. Previous models learn predicates of a single \textit{valency}, the number and types of arguments controlled by the predicate. Commonly these are binary graphs, which cannot model single-argument predicates like the entity states ``is dead'' or ``is an author.'' This means they miss a variety of entailments in text that could be used to answer questions such as our example. The Distributional Inclusion Hypothesis (DIH) \citep{dagan1999similarity, kartsaklis-sadrzadeh-2016-distributional} is a theory which has been used effectively in unsupervised learning of these same-valency entailment graphs \citep{geffet-dagan-2005-distributional, berant-etal-2010-global, hosseini2021unsupervised}.

In this work the DIH is reinterpreted in a way which supports learning entailments between predicates of different valencies such as $\textsc{kill}$(Mustard, Boddy) $\vDash \textsc{die}$(Boddy). We extend the work of \citet{hosseini2018} and develop a new \textit{Multivalent Entailment Graph} (MGraph) where vertices may be predicates of different valencies. This results in new kinds of entailments that answer a broader range of questions including entity state.

We further pose a true-false question answering task generated automatically from news text. Our model draws inferences across propositions of different valencies to answer more questions than using same-valence entailment graphs. We also compare with several baselines, including unsupervised pretrained language models, and show that our directional entailment graphs succeed over non-directional similarity measures in answering questions of fine-grained semantics. 

Advantageously, EGs are structures designed to be queried, so they are inherently explainable. This research is conducted in English, but as an unsupervised algorithm it may be applied to other languages given a parser and named entity linker.

\section{Background}
\label{sec:background}
The task of \textit{recognizing textual entailment} \cite{dagan_RTE_2005} requires models to predict a relation between a text T and hypothesis H; ``T entails H if, typically, a human reading T would infer that H is most likely true.'' From here, research has moved in several directions. We study predicates, including verbs and phrases that apply to arguments.

Research in predicate entailment graphs has evolved from ``local'' learning of entailment rules \cite{geffet-dagan-2005-distributional, szpektor-dagan-2008-learning} to later work on joint learning of ``globalized'' rules, overcoming sparsity in local graphs \cite{berant-etal-2010-global, hosseini2018}.


These graphs frequently rely on the DIH for the local learning step to learn initial predicate entailments. The DIH states that for some predicates p and q, if the contextual features of p are included in those of q, then p entails q \cite{geffet-dagan-2005-distributional}. In previous work predicate arguments are successfully used as these contextual features, but only predicates of the same valency are considered (e.g. binary predicates entail binary; unary entail unary), and further research computes additional edges in these same-valency graphs such as with link prediction \cite{hosseini2019-duality}. However, this leaves out crucial inferences that cross valencies such as the kill/die example, which are easy for humans. We generalize the DIH to learn entailments within and across valencies.

Typing is very helpful for entailment graph learning \cite{berant-etal-2010-global, lewis_semantics_2013, hosseini2018}. Inducing a type for each entity such as ``person,'' ``location,'' etc. enables generalized learning across instances and disambiguates word sense, e.g. ``running a company'' has different entailments than ``running code.''

We compare our model to several baselines, including strong pretrained language models in an unsupervised setting using similarity. BERT \cite{devlin-etal-2019-bert} generates impressive word representations, even unsupervised \citep{petroni2019language}, which we compare with on a task of predicate inference. We further test RoBERTa \cite{liu2019roberta} to show the impact of robust in-domain pretraining on the same architecture. These non-directional similarity models provide a strong baseline for evaluating directional entailment graphs.


\section{Multivalent Distributional Inclusion Hypothesis}
\label{sec:mdih}

We pose a new, multivalent interpretation of the DIH (the MDIH) which models the entailment of predicates across valencies. The intuition comes from observing eventualities \citep{Vendler-1967} which occur in the world. Neo-Davidsonian semantics \citep{davidson1967-DAVTLF, maienborn2011} explains that a textual predicate, its arguments, and adjuncts, are all properties of an underlying event variable. Entailments about one or more of the arguments arise from their roles in this eventuality. We may infer that ``Mr. Boddy died'' due to his role as a direct object in the killing/murdering event. No other information is needed, including who murdered Mr. Boddy, where, or with what instrument. Boddy is dead simply because he was murdered. We build on this insight to develop the MDIH.

Here, a predicate is represented (as in \S\ref{sec:background}) by features which are the argument tuples it appears with. We recognize a tuple as a proxy for a world event, e.g. $\textsc{visit}$(Obama, Hawaii) identifies one instance of a real \textsc{visit} event. Our method learns by tracking entity tuples across events in the world. The MDIH signals an entailment from a premise $p$ to hypothesis $h$ if, distributionally, subtuples of $p$ are always found amongst tuples of $h$. Crucially, we allow $h$ to drop in valency so that we learn entailments about subsets of $p$'s arguments. We now formalize the MDIH and then illustrate with an example. 

We define the argument tuple structures for a premise and hypothesis predicate:
\begin{align*}
    P & = \{(a_{k,1}, \ldots, a_{k,I}) \mid k \in \{1,\ldots,M\} \} \\
    H & = \{(b_{k,1}, \ldots, b_{k,J}) \mid k \in \{1,\ldots,N\} \}
\end{align*}
$P$ is a set of $M$ argument tuples (each of size $I$) which correspond to instances of a premise predicate $p$. $H$ is a set of $N$ argument tuples (each of size $J$) representing the same for hypothesis $h$. We limit $J \leq I$, e.g. we learn about relations on \textit{realized} entity subsets. We do not learn entailments to higher valencies (such as a unary entailing a binary) because additional arguments must be existential, not real. We leave this to future work.

To select argument subtuples from tuples in $P$, we define a vector of indices $\textbf{j}$ with length $J$, which selects arguments by position. For example, with $\textbf{j}=[2,3]$, perform $P[:,\textbf{j}]$. For each argument tuple in $P$, select just the 2nd and 3rd arguments, forming a new set of 2-tuples. We define the Multivalent Distributional Inclusion Hypothesis: \\

\textit{If} $P[:,\textbf{j}] \subseteq H[:,m(\textbf{j})]$, \textit{ then} $p \vDash h$ \\

Here $m: \mathbb{N}^J \rightarrow \mathbb{N}^J$ is a simple bijective mapping from argument indices of $p$ to $h$. An example where $m$ is needed for argument swapping is ``$x$ bought $y$'' entails ``$y$ sold to $x$.''

We illustrate by working the kill/die example on a hypothetical corpus. We might find that $\textsc{kill}(x, y) \vDash \textsc{die}(y)$ by trying $\textbf{j} = [2]$ and $m([2])=[1]$. 
We start with $P$, all 2-tuples of \textit{killings}, and $H$, all 1-tuples of \textit{dyings} and apply \textbf{j} and \textit{m}. We may find that selecting arg 2 from all tuples in $P$ forms a subset of the selection of arg 1 from tuples in $H$. Though \textit{dyings} may happen in many ways, we observe that arg 2 of a \textit{killing} often occurs elsewhere in the corpus with a \textit{dying}, and thus we infer the entailment between predicates. Intuitively this is true for arbitrarily large valencies: $\textsc{murder}$(Mustard, Boddy, kitchen, candlestick) entails $\textsc{kill}$(Mustard, Boddy) and both entail $\textsc{die}$(Boddy).

Though arguments may be dropped from the premise, they still influence entailments. This is because the MDIH tracks \textit{eventualities}. ``Person writing a book'' is a different kind of event than ``person writing software'' with a different distribution of argument tuples, so we learn that the former entails being an author, while the latter entails being a programmer.

\section{Learning Multivalent Graphs}
\label{sec:mgraph}

We define an Entailment Graph as a directed graph of predicates and their entailments, $G = (V, E)$. The vertices $V$ are the set of predicates, where each argument has a type from the set of 49 FIGER base types $\mathcal{T}$, e.g. $\textsc{travel.to}$(:person, :location) $\in V$, and :person, :location $\in \mathcal{T}$. The directed edges are $E = \{(v_1,v_2) \mid v_1, v_2 \in V \mbox{ if } v_1 \vDash v_2\}$, or all entailments between vertices in $V$.

In Multivalent Entailment Graphs we expand $V$ to contain predicates of both 1- and 2-valency, and $E$ to edges between these vertices, described as follows. Let $b_i, b_j \in V$ be distinct binary predicates and $u_i, u_j \in V$ be unary predicates. Define $\mathcal{E}$ as the set of all entities in the world, and some particular entities $x,y \in \mathcal{E}$ to illustrate argument slots. $E$ contains these patterns of entailment: 
\begin{enumerate}[leftmargin=*, itemsep=2pt] 
    \item $b_i{(x,y)} \vDash b_j{(x,y)}$ or $b_i{(x,y)} \vDash b_j{(y,x)}$ \\ Binary entails binary (\textbf{BB} entailments)
    \item $b_i{(x,y)} \vDash u_i{(x)}$ or $b_i{(x,y)} \vDash u_i{(y)}$ \\ Binary entails unary of one argument (\textbf{BU})
    \item $u_i{(x)} \vDash u_j{(x)}$ \\ Unary entails unary (\textbf{UU})
\end{enumerate}

Predicates with valence > 2 are sparse in the text, but are also included in the MGraph by decomposing them into binary relations between pairs of entities. This is another application of our Multivalent DIH. We maintain argument roles, so each binary is a window into its higher-valency predicate, allowing higher-valency predicates to entail lower binaries and unaries.

To learn these new kinds of connections we develop a method of local entailment rule learning using the MDIH. As in \S\ref{sec:background}, the local step learns the initial directed edges of the entailment graph, which are further improved with global learning. Our local step learns entailments by machine-reading the NewsSpike corpus (2.3GB), which contains 550K news articles, or over 20M sentences \cite{zhang-weld-2013-harvesting}. NewsSpike consists of multi-source news articles collected within a fixed timeframe, and due to these properties the articles frequently discuss the same events but phrased in different ways, providing appropriate training evidence.

\begin{figure}[h]
    \centering
    \includegraphics[width=0.55\linewidth, trim=13px 0 15px 0, clip]{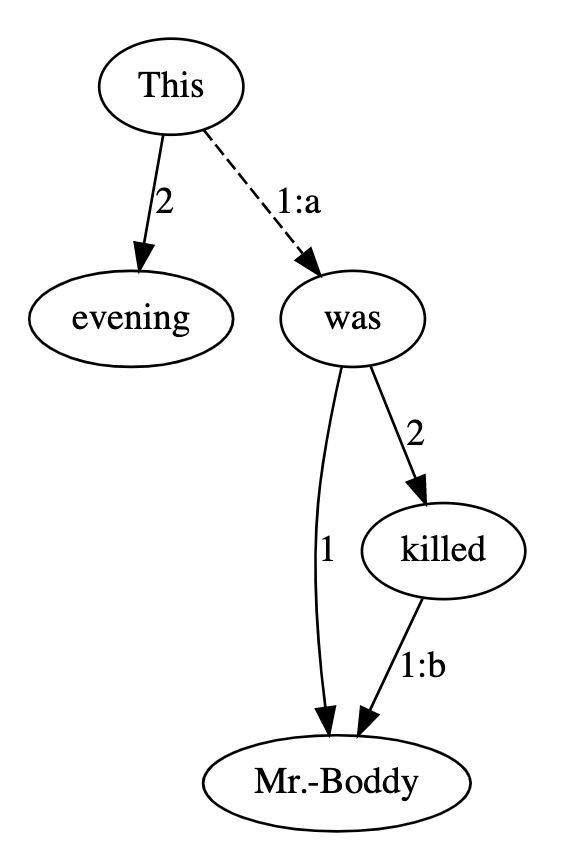} \\
    \textit{``This evening Mr. Boddy was killed.''} \\[0.5ex]
    $\Rightarrow$ \textsc{kill.2}(Mr.-Boddy)
    \caption{A sentence is CCG parsed, formed into a dependency graph (shown) using CCG dependencies, and traversed to extract a unary relation. MoNTEE traverses from a predicate to all connected arguments.}
    \label{fig:rel_extraction}
\end{figure}

\subsection{Extraction of Predicate Relations}
\label{sec:extraction}
Our pipeline processes raw article text into a list of propositions: predicates with associated typed arguments. We use the MoNTEE system \cite{devroe2021modality} to extract natural language relations between entities from raw text \footnote{We disable modality tagging in our experiments.}. This system first parses sentences using the RotatingCCG parser \cite{stanojevic-steedman-2019-ccg} (Combinatory Categorial Grammar; \citeauthor{syntactic-process}, \citeyear{syntactic-process}) and then forms dependency graphs from the parses. Finally, it traverses these graphs to extract the relations, each consisting of a predicate and its arguments. Figure~\ref{fig:rel_extraction} shows an example dependency graph and the relation extracted from it. Arguments may be either named entities\footnote{Identified by the CoreNLP Named Entitiy Recogniser \cite{CoreNLP, finkel_NER}.} or general entities (noun phrases). These entities are mapped to types by linking to their Freebase IDs \cite{bollacker-freebase} using AIDA-Light \cite{nguyen-aida-light}, and mapping the IDs to the 49 base FIGER types \cite{ling-figer}.


    

Both binary and unary relations are extracted from the corpus if they contain at least one named entity, which helps anchor to a real-world event. This poses a challenge as noted by \citet{szpektor-dagan-2008-learning}. While binary predicates may be extracted from dependency paths between two entities, unary predicates only have one endpoint, so we must carefully apply linguistic knowledge to extract meaningful unary relations. We extract these neo-Davidsonian event cases:
\begin{itemize}[leftmargin=*, itemsep=2pt]
    \item One-argument verbs including intransitives, e.g. ``Knowles sang'' $\Rightarrow \textsc{sing}$.1(Knowles) \\
    and passivized transitives, e.g. \\ ``Bill H.R. 1 was passed'' $\Rightarrow \textsc{pass}$.2(Bill-HR1)
    \item Copular constructions, where copular ``be'' acts as the main verb, e.g. \\ ``Chiang is an author'' $\Rightarrow \textsc{be.author}$.1(Chiang) \\
    and where it does not, e.g. \\  ``Phelps seems to be the winner'' $\Rightarrow \\ \textsc{seem.to.be.winner}$.1(Phelps)
\end{itemize}
As with binaries in earlier work, unary predicates are lemmatized, and tense, aspect, modality, and other auxiliaries are stripped. The CCG argument position which corresponds to its case (e.g. 1 for nominative, 2 for accusative), is appended to the predicate. 
Passive predicates are mapped to active ones. Modifiers such as negation and predicates like ``planned to'' as in ``Professor Plum planned to attend'' are also extracted in the predicate. 

We pay special attention to copular constructions, which always introduce stative predicates, rather than events \citep{Vendler-1967}. These are interesting for modeling the properties of entities.

\subsection{Learning Local Graphs}

In previous entailment graph research \citep{hosseini2018} a representation vector is computed for each typed predicate in the graph. These are compared via the DIH to establish entailment edges between predicates. The features of each vector are typically based on the argument pairs seen with that predicate. Specifically, for a typed predicate $p$ with corresponding vector $\textbf{v}$, $\textbf{v}$ consists of features $f_{i}$ which are the pointwise mutual information (PMI) of $p$ and the argument pair $a_i \in \{(e_m, e_n) \mid e_m \in \mathcal{E}_{t_1}, e_n \in \mathcal{E}_{t_2}\}$. Here $t_1, t_2 \in \mathcal{T}$, and $\mathcal{E}_t$ is the subset of entities of type $t$. For example, the predicate $\textsc{build}$(:company, :thing) might have some feature $f_{37}$, the PMI of ``build'' with argument pair (Apple, iPhone). A Balanced Inclusion (BInc) score is calculated for the directed entailment from one predicate to another \citep{szpektor-dagan-2008-learning}. BInc is the geometric mean of two subscores: a directional score, Weeds Precision \citep{weeds-2003}, measuring how much one vector's features ``cover'' the other's; and a symmetrical score, Lin Similarity \citep{lin-1998-automatic-retrieval}, which downweights infrequent predicates that cause spurious false positives.

In this work we compute local binary graphs following \citet{hosseini2018} and leverage the new MDIH to compute additional entailments for unaries and between valencies. To do this we compute a vector for each argument slot respecting its position in the predicate. For a predicate $p$, a slot vector $\textbf{v}^{(s)}$ for $s \in \{1, 2\}$ consists of features $f^{(s)}_{i}$. We define $\tau(p, s) = t$, the type of slot $s$ in predicate $p$. Each $f^{(s)}_i$ is the PMI of $p$ and the argument in slot $s$, $a^{(s)}_{i} \in \mathcal{E}_{t}$. Slot vectors are computed for the slot in unary relations and both slots in binaries. Each slot vector for $p$ has size $|\textbf{v}^{(s)}| = |\mathcal{E}_t|$, the number of entities in the data with the same type $t$.

Continuing the example, we calculate two vectors for $\textsc{build}$(:company, :thing): $\textbf{v}^{(1)} \in \mathbb{R}^{|\mathcal{E_{\text{:company}}}|}$ which contains a feature for Apple, and $\textbf{v}^{(2)} \in \mathbb{R}^{|\mathcal{E_{\text{:thing}}}|}$ which contains a feature for iPhone.

Slot vectors are comparable if they represent the same entity type. Edges are learned by comparing corresponding slot vectors between predicates. For instance, $\textsc{defeat}$(:person1, :person2) $\vDash \textsc{be.winner}$(:person1) \footnote{Here we number the typed arguments for demonstration to show which :person argument is in the entailment.} is learned by comparing the slot 1 vector of \textsc{defeat} with the slot 1 vector of \textsc{be.winner}. If the entities who have defeated someone are usually found amongst the entities who are winners then we get a high BInc score, indicating \textit{defeat} entails that its subject \textit{is a winner}.

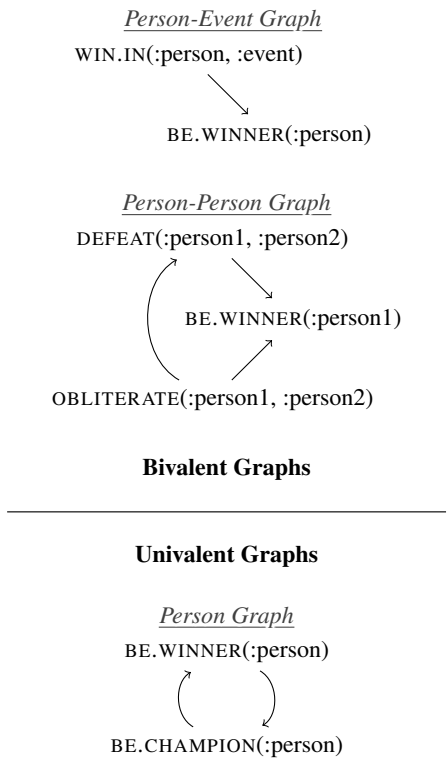
\begin{figure}[h]
\centering

\begin{minipage}{0.45\textwidth}
\centering
\small
\textcolor{darkgray}{\small \myuline{\textit{Person-Event Graph}}}\\
\begin{tikzpicture}[->,shorten >=1pt, node distance=1.5cm]
  \tikzstyle{every state}=[fill=gray,draw=none,text=white]
  
  \node (D) {\textsc{be.winner}(:person)};
  \node (E) [above left of=D] {\textsc{win.in}(:person, :event)};

  \path (E) edge (D);
  
\end{tikzpicture}
\end{minipage}
\hspace{1em} \\[0.5cm]
\begin{minipage}{0.45\textwidth}
\centering
\small
\textcolor{darkgray}{\small \myuline{\textit{Person-Person Graph}}}\\
\begin{tikzpicture}[->,shorten >=1pt, node distance=1.5cm]
  \tikzstyle{every state}=[fill=gray,draw=none,text=white]
  
  \node (A) {\textsc{be.winner}(:person1)};
  \node (B) [above left of=A] {\textsc{defeat}(:person1, :person2)};
  \node (C) [below left of=A] {\textsc{obliterate}(:person1, :person2)};

  \path (C) edge (A);
  \path (B) edge (A);
  \path (C) edge[bend left=60] (B);

\end{tikzpicture}
\end{minipage}
\\[0.5cm]
\small \textbf{Bivalent Graphs}
\\[0.15cm]
\rule{0.75\linewidth}{0.15mm}
\\[0.3cm]
\small \textbf{Univalent Graphs}
\\[0.5cm]
\begin{minipage}{0.3\textwidth}
\centering
\small
\textcolor{darkgray}{\small \myuline{\textit{Person Graph}}}\\
\begin{tikzpicture}[->,shorten >=1pt, node distance=1.25cm]
  \tikzstyle{every state}=[fill=gray,draw=none,text=white]
  
  \node (D) {\textsc{be.winner}(:person)};
  \node (E) [below of=D] {\textsc{be.champion}(:person)};

  \path (D) edge [bend left=60] (E);
  \path (E) edge [bend left=60] (D);
  
\end{tikzpicture}
\end{minipage}
\caption{Bivalent graphs model entailments from binary predicates to equal- and lower-valency predicates (binary and unary). Univalent graphs model entailments from unaries to equal-valency unary predicates.}
\label{fig:type-graphs}
\end{figure}

Figure~\ref{fig:type-graphs} illustrates a Multivalent Graph. This includes Bivalent Graphs which contain the entailments of binary predicates (BB and BU edges), and separate Univalent Graphs which contain the entailments of unary predicates (only UU edges, since we do not allow a unary to entail a binary). We follow previous research and learn separate disjoint subgraphs for each typing, up to $|\mathcal{T}|^2$ bivalent and $|\mathcal{T}|$ univalent subgraphs given enough data. For example, we learn a bivalent (:person, :location) graph containing binary predicates such as $\textsc{fly.into}$(:person, :location) which may entail unaries like $\textsc{be.airport}$(:location). 

Because a unary has only one type $t_i$ it may be entailed by binaries in up to $2*|\mathcal{T}|-1$ subgraphs with types  $\{(t_i,t_j) \mid j \in \mathcal{T}\}$, i.e. all bivalent graphs containing type $t_i$. 
We learn entailments from unaries (UU) in separate 1-type univalent graphs. This efficiently learns one set of entailments for each unary, but allows them to be freely entailed by higher-valency predicates, e.g. binaries.

Bivalent graphs point transitively into univalent graphs. In Figure~\ref{fig:type-graphs}, $\textsc{defeat}$(:person1, :person2) $\vDash \textsc{be.winner}$(:person1) in the person-person graph. E.g. further entailments of $\textsc{be.winner}$(:person) are in the person univalent graph.

\subsection{Learning Global Graphs}

Local learning of entailments suffers from sparsity issues which can be improved by further learning of ``global'' graphs. We use the soft constraint method of \citet{hosseini2018} which has two optimizations. The paraphrase resolution constraint encourages predicates within the same typed graphs that entail each other to have similar entailment patterns. The cross-graph constraint additionally encourages compatible predicates across different typed graphs to share entailment patterns.

We apply global learning to bivalent graphs and separately to univalent graphs. Globalization is valency-agnostic, using just the common structures between predicates, so bivalent graphs can use BB and BU edges to optimize binary predicate entailments. Final graph size statistics are in Table~\ref{tab:mgraph_stats}.

\begin{table}[h]
    \vspace{1em}
    \centering
    \small
    \begin{tabular}{l r r}
        \toprule
        \textbf{Valency} & \textbf{Vertices} & \textbf{Edges} \\
        \midrule
        Bivalent & 938K Binary & 94M BB / 30M BU \\[0.4em]
        Univalent & 36K Unary & 3.6M UU \\
        \bottomrule
    \end{tabular}
    \caption{We learn 546 typed bivalent subgraphs which contain entailments of binary predicate antecedents (BB and BU); and 37 typed univalent subgraphs which contain entailments of unary predicates (UU).}
    \label{tab:mgraph_stats}
\end{table}

\section{Evaluation: Question Answering}
\label{sec:qa}

We pose an automatically generated QA task to evaluate our model explicitly for directional inference between binary and unary predicates, as we are not aware of any standard datasets for this problem. Our task is to answer true-false questions about real events that are discussed in the news, for example, ``Was Biden elected?'' These types of questions are surprisingly difficult and frequently require inference to answer \cite{clark-etal-2019-boolq}. For this, entailment is especially useful: we must decide if the question (hypothesis) is true given a list of propositions from limited news text (premises), which are all likely to be phrased differently.

This task is designed independently of the MGraph as a challenge in information retrieval. Positive questions made from binary and unary predicates are selected directly from the news text using special criteria, and are then removed. From these positives we automatically generate false events to use as negatives, which are designed to mimic real, newsworthy events. The remaining news text is used to answer the questions. We attempt to make every question answerable, but since they are generated automatically there is no guarantee. However, the task is fair as all models are given the same information. The additive effects of multivalent entailment should be demonstrated: with more kinds of entailment, the MGraph should find more textual support and answer more questions.

The task is presented on a text sample from NewsCrawl, a multi-source corpus of news articles, to be published separately. A test set is extracted which contains 700K sentences from articles over a period of several months, and also a development set from a further 500K sentences. We generate questions balanced to a ratio of 50\% binary questions / 50\% unary; and within each 50\% positive / 50\% negative. Table~\ref{tab:sample_qs} shows a sample from the dev set. We generate 34,394 questions on the test set: 17,256 unary questions and 17,138 binary.

\subsection{Question Generation}
\label{sec:q_gen}
For realism, questions should be both \textit{interesting} and \textit{answerable} using the corpus. A multi-step process extracts questions from the news text itself.

\textbf{1. Partitioning.} First, the articles are grouped by publication date such that each partition covers a timespan of up to 3 consecutive days of news (49 partitions in the test set). We ask yes-no questions about events drawn from the partition, and the news text within this 3-day window is used as evidence to answer them. We ask questions as if happening presently in this time window to control for the variable of time, so we can ask ambiguous questions like ``Did the Patriots win the Superbowl?'' which may be ``true'' or not depending on the date and timespan. The small 3-day window size was chosen so multiple news stories about an event appear together, increasing the chances of finding question answers. Within each partition we do relation extraction in a process mirroring \S\ref{sec:extraction}.

\textbf{2. Selecting Positives.} We adapt a selection process from \citet{poon2009unsupervised} to choose good questions which are interesting to a human and answerable from the partition text. First, we identify repeated entities that star in the events of the articles; these will yield interesting questions as well as ample textual evidence for answering them. In each partition we count the mentions of each entity pair (from binary propositions) and single entities (from unary and binary ones). The most frequent entities and pairs mentioned more than 5 times in the partition are selected. Predicates which are mentioned across the entire news corpus 10 times or fewer are filtered out; we assume those remaining are popular to report in news and thus are interesting to a human questioner. We randomly select propositions featuring both a star entity and predicate to use as questions, and remove them from the partition.

\textbf{3. Generating Negatives.} A simple strategy for producing negatives might seem to be substituting random predicates into the positive questions. However, this is unsatisfactory because modern techniques in NLP excel at detecting unrelated words. For example, a neural model will easily distinguish a random negative like $\textsc{detonate}$(Google, YouTube) from a news text discussing Google's acquisition of YouTube, classifying it as a false event on grounds of dissimilarity alone. 

To be a meaningful test of inference this task requires that negatives be difficult to discriminate from positives: they should be semantically related but should not logically follow from what is stated in the text. To this end we derive negative questions from the selected positives using linguistic relations in WordNet \cite{wordnet}. We assume that news text follows the Gricean cooperative principle of communication \cite{sep-implicature}, such that it will report what facts are known and nothing more. To this end, noun hyponyms and their verbal equivalent, troponyms, are mined from the first sense of each positive in WordNet. For example, we extract ``burn'' as a troponym of ``hurt'' and the phrase ``inherit from'' as a troponym of ``receive from.'' We therefore expect that these specific relations will be untrue of the argument tuple in question and may be used as negatives. We also considered antonyms and other WordNet relations, but these are much sparser in English and have low coverage.

For fairness, generated negatives which actually occur in the current partition are screened out (0.1\% of proposed negatives), as well as negatives which never occur in the entire corpus (76.8\% of proposed negatives). Only challenging negatives are left, which actually do occur in real news text. See Table~\ref{tab:sample_qs} for a sample of questions. In the error analysis we find these negatives to be of good quality: they are uncommonly inferable from the text, accounting for a small percentage of false positives.

\begin{table}[h]
    \centering
    \small
    \begin{tabular}{p{0.35\linewidth} p{0.35\linewidth}}
        \toprule
        \textbf{Positive} & \textbf{Negative} \\
        \midrule
        Did the Ohio State Buckeyes \textbf{play}? & Did the Ohio State Buckeyes \textbf{fumble}? \\[0.6em]
        Was Mitt Romney a \textbf{candidate}? & Was Mitt Romney a \textbf{write-in}? \\[0.6em]
        \raggedright Did voters \textbf{reject} Mike Huckabee? & Did voters \textbf{discredit} Mike Huckabee? \\[0.4em]
        \raggedright Did Roger Clemens \textbf{receive from} Brian McNamee? & \raggedright Did Roger Clemens \textbf{inherit from} Brian McNamee? \tabularnewline 
        \bottomrule
    \end{tabular}
    \caption{A sample of dev set questions.}
    \label{tab:sample_qs}
\end{table}

\subsection{Question Answering Models}

In each partition, models receive factual propositions extracted from 3 days of news text to use as evidence for answering true-false questions. A model scores how strongly it can infer the question proposition from each evidence proposition, and we take the maximum score as the model confidence of a ``true'' answer.

\textbf{Exact-Match.} Our text is multi-source news articles, so world events are often discussed multiple times in the data, even with the same phrasing. We compute an ``exact-match'' baseline which shows how many questions can be answered from an exact string match in the text; the rest require inference.

\textbf{Binary Entailment Graph.} Our BB model is roughly equivalent to the state of the art binary-to-binary entailment graph \citep{hosseini2018}, so it serves as a baseline for the overall model. \footnote{We test the MGraph on the Levy/Holt dataset of 18,407 questions for BB entailment \cite{levy-dagan-2016-annotating, holt_2018}, and achieve similar results to \citet{hosseini2018}.}

All graph models look for directed entailments from evidence propositions to the question proposition. For example, ``Was YouTube sold to Google?'' can be answered affirmatively by reading ``Google bought YouTube'' using the graph edge $\textsc{buy}(x,y) \vDash \textsc{sell.to}(y,x)$. BInc scores range from 0 to 1; if no entailments are found we assume it is false (score of 0).


\textbf{Multivalent Entailment Graph.} The MGraph is made of 3 component models: (1) the BB model which uses binary evidence to answer binary questions; (2) the UU model which uses unary evidence to answer unary questions; and (3) the BU model which uses binary evidence to answer unary questions. The MGraph is able to answer questions using evidence across valencies, e.g. ``Is J.K. Rowling an author?'' is affirmed by reading ``J.K. Rowling wrote \textit{The Sorcerer's Stone}'' using the graph edge $\textsc{write}(x,y) \vDash \textsc{be.author}(x)$. Individually, each model answers only binary or unary qustions, not both. By combining them all kinds of questions can be answered using all available evidence. At each precision level if any component model predicts true, the overall model does too.


In some test instances the entity typer may make an error, and so we fail to find the question predicate in the typed subgraph. Similarly to \citet{hosseini2018}, in these cases we back off, querying all subgraphs for the untyped predicate and averaging the entailment scores found. We find 5\% more unary questions and  18\% more binaries.

\begin{figure*}[h]
    \centering
    \includegraphics[width=0.99\textwidth]{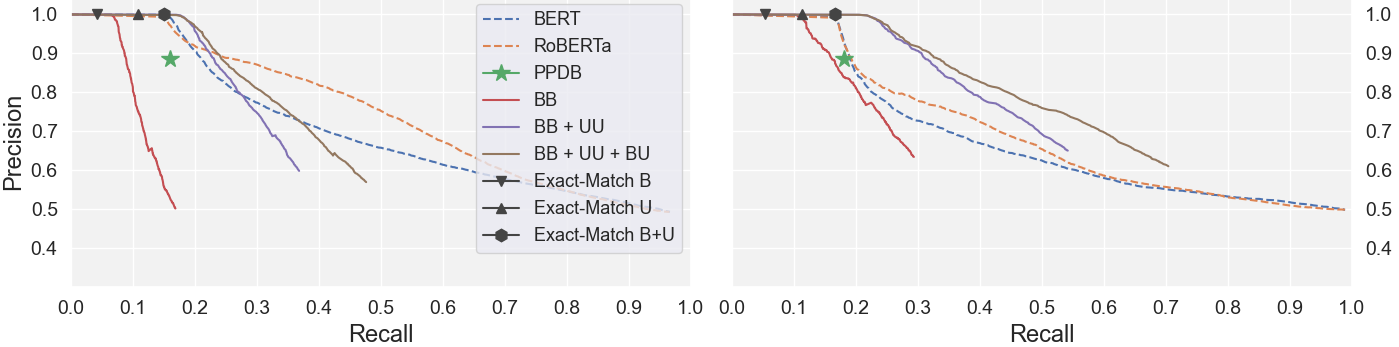}
    \caption{(Left) Overall performance on the QA task. (Right) performance on the filtered task. Note that BB, UU, and BU models may individually reach a max recall of 50\% because they answer only binary or unary questions.}
    \label{fig:all_performance}
\end{figure*}

\textbf{Similarity Models.} BERT and RoBERTa predicate embeddings \cite{devlin-etal-2019-bert, liu2019roberta} are used in an unsupervised manner to answer questions based on similarity to the evidence. We encode the question into a representation vector, and each evidence proposition with the same arguments. We compute the cosine similarity between the question and each evidence vector, adjusted to a scale of 0 to 1: $\text{sim}(\textbf{p}, \textbf{q}) = (cos(\textbf{p},\textbf{q})+1)/2$. 

To compute each vector encoding we construct a simple natural language sentence from the proposition using its predicate and arguments 
and encode it with the language model. Our representation includes \textit{only} the encoding for the predicate in the context of its arguments, but not the arguments themselves to make this a true test of predicate similarity. We average all final hidden-state vectors from the model corresponding to the predicate, excluding those of the arguments. We test the basic BERT model and RoBERTa model, which has robustly pretrained on 160GB  of text (76GB news).

\textbf{PPDB.} Though supervised, PPDB 2.0 (we use XXXL) \cite{pavlick-etal-2015-ppdb} is a useful comparison as it is a large, well-understood resource for phrasal entailment. PPDB relations come from bilingual pivoting and are categorized using text-based features, which is very different from our argument-tracking method. We view PPDB as a kind of Entailment Graph with 9M predicate phrases (vertices) and 33M ``Equivalence'' and ``ForwardEntailment'' edges. We convert evidence and question propositions into a natural text format and extract a PPDB relation score from each evidence phrase to the question. 

\section{Question Answering Results}
The models produce a gradation of judgement scores between 0 (false) and 1 (true). As in earlier work, we slide a classification threshold over the score range to produce a precision-recall curve for each model. Results are in Figure~\ref{fig:all_performance} (left).


Multivalent graph performance is shown incrementally. The BB model can answer a portion of binary questions; the UU model can answer more unary questions; adding the BU model can answer still more unary questions using binary evidence. We observe successful inference of our kill/die example and others. ``Obama was elected to office'' affirms the question ``Was Obama a candidate?'' and ``Zach Randolph returned'' affirms ``Did Zach Randolph arrive?''

Our test set is from multiple sources over the same time period. The exact-match baseline shows the limitations of answering questions simply by collecting more data; most questions require inference to answer. The complete MGraph achieves \textasciitilde 3x this recall by drawing inferences.

Our model achieves higher precision than BERT and RoBERTa similarity models in the low recall range. The similarity models perform well, achieving full recall by generalizing for rarer predicates. We note that RoBERTa bests BERT due to extensive in-domain pretraining.


The BB model appears to struggle. In fact 90.5\% of unary questions have a vertex in the graph, but only 64.1\% of binaries do. The BB model frequently cannot answer questions because the question predicate wasn't seen in training. This difference is because binary predicates are more diverse so suffer more from sparsity: they are often multi-word expressions and have a second, typed argument. Indeed, most binary predicate research (in symbolic methods) focuses on only the top 50\% of recall in several datasets \cite{berant-etal-2010-global, berant-etal-2015-efficient, levy-dagan-2016-annotating, hosseini2018}.

\begin{table}[h]
\small
\centering
\begin{tabular}{lllll}
\toprule
& \multicolumn{2}{l}{\bf Unary Questions} & \multicolumn{2}{l}{\bf Binary Questions} \\
\cmidrule(lr){2-3} \cmidrule(lr){4-5}
\textbf{Model} & @1451 & @2000 & @802 & @2000 \\
\midrule
BERT & 91.4\% & 76.9\% & 92.0\% & 82.9\% \\
RoBERTa & 92.5\% & 78.6\% & 91.5\% & 85.1\% \\
PPDB    & 92.3\% & — & 81.8\% & — \\
\midrule
\textit{MGraph} \\
\quad UU & 96.5\% & 87.0\% & — & — \\
\quad BU & 97.6\% & 90.4\% & — & — \\
\quad BB & — & — & 100.0\% & 88.8\% \\
\midrule
    & \multicolumn{2}{l}{1245 Exact-Match} & \multicolumn{2}{l}{597 Exact-Match} \\
\bottomrule
\end{tabular}
\caption{The filtered test. Models rank question/answer pairs by confidence. We show accuracy on the $K$ most confident predictions, at two points. PPDB doesn't answer enough questions to reach the @2000 cutoff, so we also compare at the smaller PPDB maximum.}
\label{tab:breakdown}
\end{table}

For an even comparison we create a filtered question set. From all questions we remove those without a vertex in the MGraph, then balance them as in \S\ref{sec:qa}, resulting in 20,519 questions (10,273 unary and 10,246 binary). This filtered test directly compares the models, since both the entailment graphs and the similarity models have a chance to answer all the questions. Results are shown in Figure~\ref{fig:all_performance} (right), with a very different outcome. Head-to-head, the MGraph offers substantially better precision across all recall levels. At 50\% recall, the MGraph has 76\% precision with RoBERTa at 65\%.


Notably, on both tests, more unary questions are answered using both unary \textit{and} binary predicate evidence than just using unary evidence alone. On the filtered test, the BU model increases max recall from 54\% to 70\%.

Finally, we note PPDB's poor performance (highest recall shown), only 1\% higher recall than the exact-match baseline despite having entries for 88\% of questions. Though PPDB features many directional entailments, this sparsity of edges useful for the task is likely because bilingual pivoting excels at detecting near-paraphrases, not relations between distinct eventualities, e.g. it can't learn ``getting elected'' entails ``being a candidate.'' Advantageously, our method learns this open-domain knowledge by tracking entities across all the events they participate in.

We show a breakdown of the filtered test results in Table~\ref{tab:breakdown}. Models don't answer all the questions, so following \citet{lewis_semantics_2013} who design a similar QA task, we evaluate models on the accuracy of their $K$ most confident predictions.

\section{Error Analysis}

We sample 300 false positives (100 for each model) and report analyses in Table~\ref{tab:error_analysis}. In all models spurious entailments are the largest issue, and may occur due to normalization of predicates during learning, or incidental correlations in the data. The UU and BU models also suffer during relation extraction (parsing). When we fail to parse a second argument for a predicate we assume it only has one and extract a malformed unary, which can interfere with question answering (e.g. reporting verbs ``explain,'' ``announce,'' etc. which fail to parse with their quote). We also find relatively few poorly generated negatives, which are actually true given the text. In these cases the model finds an entailment which the authors judge to be correct.



\section{Conclusions}

The MDIH is shown as an effective theory of unsupervised, open-domain predicate entailment, which crosses valencies by respecting argument roles.

Our multivalent entailment graph's performance has been demonstrated on a question answering task requiring fine-grained semantic understanding. Our method is able to answer a broader variety of questions than earlier entailment graphs, aided by drawing on evidence across valencies. We outperform baseline models including a strong similarity measure using unsupervised BERT and RoBERTa, while using far less training data. This shows that directional entailment is more helpful for inference on such a task than non-directional similarity, even with robust, in-domain pretraining.

We also noted a complementarity between unsupervised methods. Our symbolic graph method achieves high precision for learned predicates, while sub-symbolic neural models achieve high recall by generalizing to unseen predicates. Future work may leverage our MDIH signal to train a directional neural classifier and combine benefits.

\begin{table}[t]
    \small
    \begin{tabular}{p{0.35\linewidth} p{0.53\linewidth}}
        \toprule
        \textbf{Error Source} & \textbf{False Positive Example} \\
        \midrule
        \multicolumn{2}{c}{\bf Unary to Unary (UU) Judgements} \\
        \midrule
        Spurious Entailment (57\%) & The United States advances $\vDash$ The United States falls \\[0.3em]
        Parsing (26\%) & Reuters reports $\vDash$ Reuters notes \\[0.3em]
        \raggedright Poor Negative (actually true) (17\%) & Productivity increases $\vDash$ Productivity grows \\
        \midrule
        \multicolumn{2}{c}{\bf Binary to Unary (BU) Judgements} \\
        \midrule
        Spurious Entailment (65\%) & New York Mets create through camerawork $\vDash$ New York Mets benefit \\[0.3em]
        Parsing (26\%) & John McCain spent part of 5 years $\vDash$ John McCain drew \\[0.3em]
        \raggedright Poor Negative (actually true) (9\%) & \raggedright The Yankees overwhelm the Mariners $\vDash$ the Yankees prevail \tabularnewline
        \midrule
        \multicolumn{2}{c}{\bf Binary to Binary (BB) Judgements} \\
        \midrule
        Spurious Entailment (53\%) & A soldier was killed in Iraq $\vDash$ A soldier was murdered in Iraq \\[0.3em] 
        \raggedright Poor Negative (actually true) (32\%) & Profits fall in the first quarter $\vDash$ Profits decline in the first quarter \\[0.3em]
        Parsing (17\%) & \raggedright medal than United States $\vDash$ United States take the medal \tabularnewline 
        \bottomrule
    \end{tabular}
    \caption{False positive analysis. Models predict entailments from the text (left) to generated negatives (right).}
    \label{tab:error_analysis}
\end{table}

\section*{Acknowledgments}
This work was supported in part by ERC H2020 Advanced Fellowship GA 742137 SEMANTAX, and an Edinburgh and Huawei Technologies Research Centre award.

\bibliography{anthology,custom,sources}
\bibliographystyle{acl_natbib}

\end{document}